\documentclass{article}
\usepackage{spconf,amsmath,graphicx}
\usepackage{times}
\usepackage{latexsym}
\usepackage{dirtytalk}
\usepackage{graphicx}
\usepackage{float}
\usepackage{url}
\usepackage{amsmath}
\usepackage[linesnumbered,ruled]{algorithm2e}
\usepackage[T1]{fontenc}
\usepackage{verbatim}
\usepackage{multirow}

\title{LEARNING TO MIRROR SPEAKING STYLES INCREMENTALLY}
\name{Siyi Liu$^{\star}$ \qquad Ziang Leng$^{\star}$ \qquad Derry Wijaya\thanks{$^{\star}$ Contributed equally}}

\address{Boston University}
\begin{document}
%
\maketitle
\begin{abstract}
  Mirroring is the behavior in which one person subconsciously imitates the gesture, speech pattern, or attitude of another \cite{chartrand1999chameleon}. In conversations, mirroring often signals the speakers' enjoyment and engagement in their communication. In chatbots, methods have been proposed to add personas to the chatbots and to train them to speak or to shift their dialogue style to that of the personas. However, they often require a large dataset consisting of dialogues of the target personalities to train. In this work, we explore a method that can learn to mirror the speaking styles of a person incrementally. Our method extracts n-grams that capture a person's speaking styles and uses the n-grams to create patterns for transforming sentences to the person's speaking styles. Our experiments show that our method is able to capture patterns of speaking style that can be used to transform regular sentences into sentences with the target style. 
\end{abstract}
\section{Introduction}
With the growth of deep learning, natural language processing, human-computer interaction, and dialogue research, the development of chit-chat models and other conversational agents with end-to-end neural approaches has become very popular. A popular class of models are generative recurrent neural network models such as \textsc{seq2seq} applied to dialogue. However, since these models require training on a large dataset, which contain many dialogues with different speakers, they lack a consistent personality \cite{li2016persona}. 

To make user interactions with chatbots more human-like and engaging, methods have been proposed to add persona to the models by encoding speaker-level vector representations into a \textsc{seq2seq} model \cite{li2016persona} or by training the model on human-human conversation data where the interlocutors act the part of a given provided persona \cite{zhang2018personalizing}, thus training the model to mimic and generate persona-specific responses. These persona however, only model the human-like character or personal topics (i.e., the ``profile") of the speakers rather than their specific linguistic styles. 

Separately, other methods have been proposed to transform chatbot responses to mimic linguistic styles of specific domains such as Star Trek \cite{jena2017enterprise} or entertainment and fashion \cite{banerjee2016transforming}. These methods are trained on datasets of conversations involving many speakers from the domains and thus do not model individual speaker styles. 

In this work, we propose an incremental method for capturing an individual speaking style and for transforming regular sentences into sentences of that style. Just as humans would mirror their conversation partner's speech pattern as the conversation progresses, our model learns to capture the style of a person's talking incrementally, updating its patterns as it receives more conversation data. By proposing a method that can learn from a growing data, our method is different from all previous methods that require large training data. 

Our method captures speech patterns by extracting frequently occurring n-grams from a speaker's speech based on efficient algorithms for frequent itemset mining \cite{agrawal1993mining,srikant1996mining}.  
Our hypothesis is that the frequent n-grams in a speaker's speech capture his linguistic style while the remainder of his sentences captures context.
\begin{figure}
  \includegraphics[width=\linewidth]{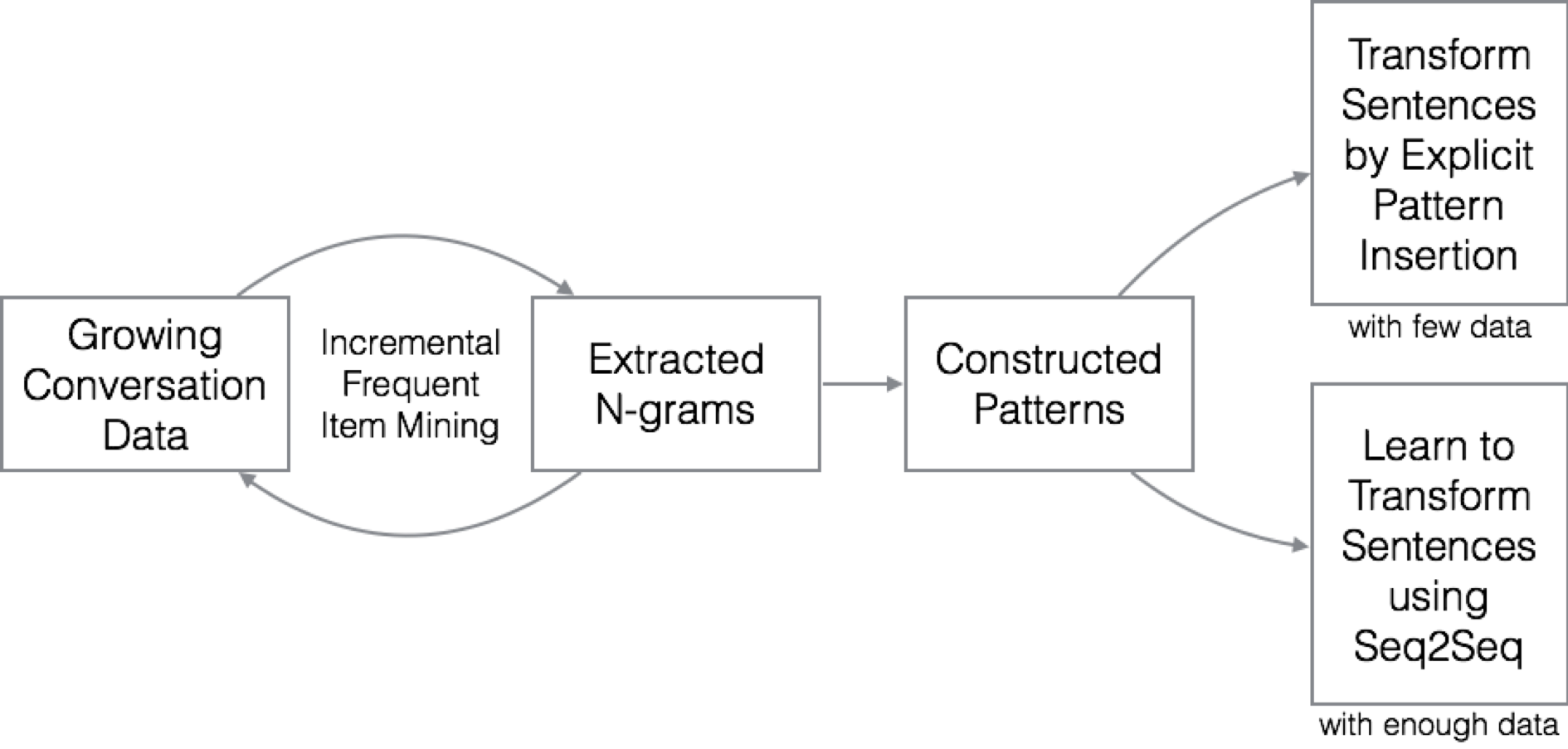}
  \caption{Methods we use to mirror style under different resource settings}
  \label{fig:boat1}
\end{figure}
The patterns are then used to transform sentences into those of the speaker's style using two methods: 
(1) by using a state-of-the-art language model to construct vector representations of the sentences and cosine similarities of these vectors, to select the best pattern to insert into the input sentence, (2) by training a \textsc{seq2seq} network to select the pattern \textit{and} transform the input sentence jointly. We conjecture that our first method will work better when there is few data (e.g., at the beginning of conversing with the speaker) while our second method will work better when we have a large enough speaker's data (Figure~\ref{fig:boat1}) and experiment with varying amount of data. We observe that our explicit pattern insertion approach works better at transforming sentences than our neural network approach given few data. 

We foresee a variety of applications for our incremental modeling of a user's speech pattern. 
Our method can be used to transform chatbot responses to mirror the user's style as the conversation progresses (just as humans would) to relate better to the user. It can also be combined with the persona-based models that encode the user's profile, thus generating responses that are personal both in terms of context \textit{and} style. As users may use different linguistic styles when speaking to different social groups (e.g., a student may use different styles when talking to friends as opposed to his professors), by learning to capture these styles we can generate more natural chatbot responses to users depending on the social groups.

\section{Related work}
In mirroring user's linguistic styles, the closest approach to ours is the work of \cite{banerjee2016transforming} that describes how to transform a sentence into a specific domain's style of speaking. The motivation being that a user interested in politics or entertainment domain might enjoy getting chatbot responses that resemble the speaking styles of politicians or entertainers, respectively. The method proposes to construct word-graph based on tweets published by several famous personalities from these domains, 
where an edge is created between two words in the word-graph if they are adjacent in the tweet. 
To transform an input sentence into a specific domain's style, they insert a relevant path (i.e., word sequences) in the word-graph between each pair of words in the input sentence following a set of manually defined rules. 
Finally, word sequences inserted between each pair of words in the input sentence are combined to generate candidate transformed sentences, which are then ranked based on their cosine similarities to the input sentence. This method is used to transform chatbot responses to a specific domain's style, for example to transform regular \textsc{seq2seq}-generated chatbot responses in Star Trek style \cite{jena2017enterprise}. 

These works are different from our work in that they model domain's specific style instead of an individual's specific style and that they are static methods instead of incremental. Although one can envision building a word-graph that can be incrementally updated as new data comes in, these works do not experiment with varying amount of data for constructing the word-graph or assess if the method could scale -- as constructing a word-graph using adjacency relations can result in a large number of edges. Finally, instead of using manually defined heuristics to identify patterns of word usages efficiently, we formalize our approach based on frequent itemset mining algorithms \cite{agrawal1993mining, srikant1996mining}, which are established algorithms for finding frequent items in a dataset that have the benefit of being efficient (scale to Web-scale corpora \cite{nakashole2012patty}) and being incremental (can deal with a growing dataset). Specifically, in its incremental version \cite{thomas1997efficient}, the algorithm keeps track of the frequent items and items with the potential of being frequent in an incremental dataset. Hence, it can mine incoming stream of data without having to re-mine previous data.  

\section{Method}
We start by extracting frequent n-grams from a person's speech. Our hypothesis is that these frequent n-grams capture the person's speaking style. The extracted n-grams are used to break down the sentences in the person's speech into wildcard-separated subsequences, which we treat as patterns of the person's speech. Given an input sentence, we use the patterns to transform the input sentence into a sentence with the person's speaking style using two different methods depending on the availability of training data (i.e., sentences spoken by the person). When there is few data, we use BERT language model \cite{devlin2018bert} (whose vector representations have obtained state-of-the-art results on a wide array of Natural Language Processing tasks) to construct vector representations of the sentences. Based on these vector representations, we select a pattern whose originating sentences (with the pattern removed) have the highest cosine similarity with the input sentence. We then transform the input sentence into a sentence with the person's speaking style by explicitly injecting the selected pattern into the sentence. 

In the second method, when there is enough data, we train a \textsc{seq2seq} network with attention \cite{bahdanau2014neural} -- popularly used in neural machine translation -- to encode the input sentence and decode an output sentence with the target speaking style. We train the network on input-output pairs of sentences constructed from the person's speech. The inputs are the sentences' contexts (i.e., sentences with the frequent n-gram subsequences removed) and the outputs are the corresponding original sentences. In the following, we detail the various steps of our method. 


\subsection{Frequent N-gram Extraction}
Our frequent n-gram extraction algorithm looks for the frequent \textit{n} consecutive words (i.e., n-grams) in sentences from a person's speech. Similar to the idea of association rule mining \cite{agrawal1993mining, srikant1996mining}, we set up a minimum support as a threshold and if any n-gram's support: the proportion of sentences in the person's speech that contains the n-gram is more than this minimum support, 
we will take it as a frequent n-gram. The algorithm starts by finding frequent \textit{uni}-grams all the way up to \textit{n}-grams until it cannot find any (\textit{n+1})-gram occurring more than the minimum support. Specifically, we utilize the the downward-closure property of support in association rule mining \cite{agrawal1993mining}, which guarantees that for a frequent n-gram, all its subsequences are also frequent and thus no infrequent m-gram (m $<$ n) can be a subset of a frequent n-gram. 
Our n-gram extraction method first looks for frequent uni-grams, and based on sentences where these uni-grams occur, looks for frequent bi-grams in these sentences, and so on. After several iterations, we have all frequent n-grams of the person's speech. We filter the frequent n-grams and remove n-grams that consist only of stop words. This filters out n-grams such as ``it was" or ``she is" which are general and not specific to any person's speech. 

To efficiently extract n-grams from a growing conversation data, we use an incremental version of the association rule mining algorithm \cite{thomas1997efficient} whose key idea is to maintain two lists of n-grams, one that contains frequent n-grams found so far and another that contains potentially frequent n-grams: n-grams that are not frequent but whose subsequences are all frequent. These potentially frequent n-grams can be promoted to frequent n-grams with more data. By maintaining the two lists, the algorithm only needs to mine the new data without having to re-calculate supports from previous data. Once the algorithm extracts all frequent n-grams from the data so far, we use the same n-gram filtering as before. 

\subsection{Pattern Construction}
\label{sect:pdf}
To generate patterns of a person's speaking style from the sentences in his speech and the frequent n-grams extracted from them, we decompose the sentences into \textit{patterns}, which contain n-grams that appear frequently in the person's speech with the remaining word sequences replaced by wildcards. 
Specifically, for each sentence, we iterate through the extracted frequent n-grams (from longest n-grams to bi-grams, as longer n-grams are more indicative of one's speaking style) 
and preserve all the frequent n-grams that this sentence contains while changing the rest of the sentence to \say{*}. 

For example, the sentence \say{I will try my best to bring our jobs back} will result in the \textit{pattern} \textit{* try my best to *}, if ``try my best to" is one of the frequent n-grams extracted from the person's speech. The remainder of the sentence (i.e., word sequences that are replaced by \say{*}, which in this example is: ``I will bring our jobs back") is treated as one of the \textit{contexts} in which this pattern appears. 
After iterating through the person's speech, we will have gathered a list of patterns that capture the person's speaking style. For each pattern $p$, we maintain its list of contexts $C_p$ and the list of original sentences $O_p$ that the pattern appears in, which in the previous example is: ``I will try my best to bring our jobs back".

We use the patterns, their contexts, and their original sentences to transform an input sentence into a sentence of a specific style by explicitly injecting a pattern into the input sentence \textit{or} by using our \textsc{seq2seq} network. The key idea of both methods is to maintain the \textit{context} of the sentence (i.e., the transformed sentence should be \textit{contextually} similar to the input sentence) while \textit{re}-writing the sentence using a \textit{pattern}. 




\subsection{Sentence Transformation with Few Data}
Neural network approaches often do not work well when trained with few data. To transform sentences when there is few available data to train on, we utilize BERT's pre-trained language model \cite{devlin2018bert} to transfer learn from its large training corpora and construct vector representation of sentences from a person's speech. 

Given an input sentence, we select the best pattern for the input sentence by computing for each pattern $p$, the cosine similarity between the vector representation of the input sentence and the average vector representation of the pattern's contexts $C_p$. The pattern with the highest cosine similarity is selected and explicitly injected into the input sentence. To do so, we first split the input sentence into consecutive n-grams using chunking and replace the wildcards (\say{*}) in the pattern with these n-grams to generate candidate transformed sentences. 
For example, given an input sentence ``I eat an instant noodle", if the selected pattern is \textit{* try my best to *}, the candidate transformed sentences include ``try my best to I eat an instant noodle", ``I try my best to eat an instant noodle", ``I eat try my best to an instant noodle", and ``I eat an instant noodle try my best to".
We compare the cosine similarity between each candidate sentence vector to the average vector representation of the original sentences of the pattern $O_p$ and output the most similar sentence as our transformed sentence.

To ensure that the output sentence is grammatical, we input the sentence into a state-of-the-art grammar error correction (GEC) system \cite{zhao2019improving} that performs GEC by copying the unchanged words and the out-of-vocabulary words directly from the source sentence to the target sentence and by pre-training on the unlabeled One Billion Benchmark dataset \cite{chelba2013one}. 

\subsection{Sentence Transformation with Large Data}
\label{ssec:layout}
To learn to transform an input sentence into a sentence with a particular style, we train a \textsc{seq2seq} network with attention \cite{bahdanau2014neural} using sequences in patterns' contexts $C_p$ as inputs and the original sentences $O_p$ as outputs for all patterns $p$ we extracted from the person's speech. The network is thus trained to transform a sentence context \textit{without} the (n-grams) pattern to a sentence \textit{with} the pattern while maintaining context. 

The trained network can then take an input sentence and transform it to a sentence with the person's pattern. The \textsc{seq2seq} encoder-decoder is trained to encode the input sentence such that sentences with similar contexts are close in the encoding space and are decoded to output sentences of similar styles. With a large enough data, we believe the \textsc{seq2seq} network will perform better at transforming sentences than the previous method. 

\section{Experiment, Results, and Discussion}
We apply our proposed methods on different amount (5\%, 10\%, 20\%, to all) of the 17k sentences in Donald Trump's 2016 campaign speeches  
to capture his style of speaking and transform regular sentences into that style. 

To obtain a sentence vector using BERT, we perform average pooling on the second to the last hidden layer of all tokens in the sentence from BERT's pre-trained model (base, uncased) and use it as the sentence embedding. To extract frequent n-grams, we experiment with different values for the minimum support and report experiment results for the minimum support of 0.6\%, which result in relatively fewer (<30) patterns. 
We clean sentences in our dataset by lower casing them and by removing special characters and punctuation. We use vanilla \textsc{seq2seq} model with Bahdanau attention and GRU
. We trained the \textsc{seq2seq} model with 5\%, 10\%, 20\%, and full dataset with 40, 30, 20, and 20 epochs respectively.



Using the patterns, their contexts, and original sentences extracted from Donald Trump's speeches, we train the \textsc{seq2seq} network and transform input sentences into sentences with Trump's style. Some of the transformed sentences are shown in Table \ref{tab:nn}. As can be seen, the transformed sentences exhibit Donald Trump's style of speaking such as his use of fillers and simple words found in casual conversations like ``\textit{you know}", ``\textit{I said}", or ``\textit{I mean}" or his repeated use of intensifying adjectives and adverbs like ``\textit{best}", ``\textit{again}", which have been widely discussed by linguists and cognitive scientists \cite{npr}.

\begin{table}[!htbp]
    \small
    \centering
    \begin{tabular}{|p{4cm}|p{3.5cm}|}
        \hline
        \textbf{Input Sentence} & \textbf{Transformed Sentence} \\
        \hline
        we're going to keep winning & I mean we're going to keep winning again  \\
        \hline
        we have crowds like this everywhere no matter where we go & we have the best crowds like this \\
        \hline
        take a look at this crowd & i said take a look at this crowd \\
        \hline
        he's never seen anything like it & and you know he's never seen anything like it \\
        \hline
        tough & we need tough \\
        \hline
    \end{tabular}
    \caption{Examples of sentences transformed using \textsc{seq2seq} trained with all 17k sentence pairs}
    \label{tab:nn}
\end{table}

We also show examples of differences between our two methods (BERT and explicit pattern injection and \textsc{seq2seq}) for transforming sentences when trained on the same 5\%, 10\%, and 20\% subset of the data in Table \ref{tab:comparison}. As we hypothesized, the neural network approach does not work very well at transforming sentences when there is only few data to train with, resulting in rambling and incoherent sentences with repeated words that are unrelated to the input. BERT with explicit pattern injection works better at transforming the sentences into Trump's style while maintaining the input context. 

\begin{table}[h]
\small
\begin{tabular}{|p{0.5cm}|p{2cm}|p{2.5cm}|p{2cm}|}
\hline
\textbf{\%} & \textbf{Input} & \textbf{BERT+Explicit Pattern} & \textbf{\textsc{seq2seq}} \\
\hline
5\% & and i put up millions & and i put up millions \textbf{all over the place} & and we don't know what we don't know what we don't know what ...\\
\hline
10\% & those people have followed me all over the place & \textbf{look at }those people who have followed me all over the place & I love the world\\
\hline
20\% & we send them wheat & \textbf{you know} we sent them wheat & I love the world is going to be great again\\
\hline
\end{tabular}
\caption{Examples of sentences transformed using 5\%, 10\%, and 20\% of the data using explicit pattern injection (bolded) and \textsc{seq2seq}}
\label{tab:comparison}
\end{table}
\begin{table}[!htbp]
    \small
    \centering
    \begin{tabular}{|c|c|c|c|c}
        \hline
        \textbf{Metric}& \textbf{\%} & \textbf{BERT+} & \textbf{\textsc{seq2seq}}\\
        & \textbf{} & \textbf{Explicit Pattern} & \\
        \hline
        \multirow{4}{*}{Perplexity} & 5\% & 27.5 & 11.7 \\
        
        &10\%& 22.5 & 27.0\\
        
        &20\%&22.5&24.2\\
        &100\%& -- & 36.7 \\
        \hline
        \multirow{4}{*}{Cosine-Similarity} & 5\% & 0.98 & 0.68\\
        
         & 10\% & 0.98 & 0.72\\
        
         & 20\% & 0.98 & 0.74\\
         &100\% & -- & 0.85\\
        \hline
    \end{tabular}
    \caption{Median perplexity of transformed sentences and their average similarities to input sentences}
    \label{tab:metric}
\end{table}
We use a word-level neural language model trained on Trump's speech to measure the perplexity of the models' transformed sentences as a proxy for their performance at transforming sentences into Trump's style. We use cosine similarities between weighted average word vectors \cite{arora2016simple} of the input and the models' transformed sentences  as a proxy for their ability to maintain the input context.

As shown in Table~\ref{tab:metric}, sentences generated by \textsc{seq2seq} when there is few data ($\leq$ 20\%) have artificially low perplexities as they contain either exact or repeating sentences from the training data which may be fluent but are unrelated to the input (seen from the low similarities to the input sentence). This aligns with findings on Neural Machine Translation (NMT) \textsc{seq2seq} models, which have been observed to produce fluent but unrelated output when trained with few data \cite{koehn2017neural}. Given the same few data however, sentences generated using BERT and explicit pattern injection can effectively reflect the speaking patterns of Trump in terms of his common usage of words (low perplexity) and while maintaining input context (seen from the high similarities to the input sentence). 
In conclusion, in this paper, we presented a method for capturing a speaker's linguistic style and to transform sentences into that style. Our method can be applied to a growing speaker's data, simulating the way humans are able to adapt to and mirror their speaking partner's style as their conversation progresses. We present an incremental method for capturing the speaker's style and two methods for transforming sentences into that style that perform well both when there is few data (BERT+Explicit Pattern) and when there is a large enough data (\textsc{seq2seq}). 

\bibliographystyle{IEEEbib}
\bibliography{strings,refs}

\end{document}